\pdfoutput=1

\documentclass[11pt]{article}

\usepackage[final]{coling}

\usepackage{times}
\usepackage{latexsym}

\usepackage[T1]{fontenc}

\usepackage[utf8]{inputenc}

\usepackage{microtype}

\usepackage{inconsolata}

\usepackage{graphicx}
\usepackage{algorithm}
\usepackage{algorithmic}
\usepackage{todonotes}
\usepackage{verbatim}
\usepackage{adjustbox}
\usepackage{listings}
\usepackage[]{mdframed}
\usepackage{tabularx}
\usepackage{array}
\usepackage{todonotes}
\usepackage[normalem]{ulem}
\usepackage{framed}
\usepackage{tcolorbox}

\title{Hybrid-SQuAD: \textbf{Hybrid} \textbf{S}cholarly \textbf{Qu}estion \textbf{A}nswering \textbf{D}ataset}

\author{
 \textbf{Tilahun Abedissa Taffa\textsuperscript{1,2}},
 \textbf{Debayan Banerjee\textsuperscript{2}},
 \textbf{Yaregal Assabie\textsuperscript{3}}, \and
 \textbf{Ricardo Usbeck\textsuperscript{2}}
\\
\\
 \textsuperscript{1}Semantic Systems, University of Hamburg, Hamburg, Germany \\
 \textsuperscript{2}Artificial Intelligence and Explainability, Leuphana University Lüneburg, Lüneburg, Germany\\
 \textsuperscript{3}Department of Computer Science, Addis Ababa University, Addis Ababa, Ethiopia
\\
 \small{
   \textbf{Correspondence:} \href{mailto:email@domain}{tilahun.taffa@uni-hamburg.de}
 }
}

\begin{document}
\maketitle
\begin{abstract}
Existing Scholarly Question Answering (QA) methods typically target homogeneous data sources, relying solely on either text or Knowledge Graphs (KGs). However, scholarly information often spans heterogeneous sources, necessitating the development of QA systems that can integrate information from multiple heterogeneous data sources. To address this challenge, we introduce Hybrid-SQuAD (\textbf{Hybrid} \textbf{S}cholarly \textbf{Qu}estion \textbf{A}nswering \textbf{D}ataset), a novel large-scale QA dataset designed to facilitate answering questions incorporating both text and KG facts. The dataset consists of 10.5K question-answer pairs generated by a large language model, leveraging the KGs - DBLP and SemOpenAlex alongside corresponding text from Wikipedia. In addition, we propose a RAG-based baseline hybrid QA model, achieving an exact match score of 69.65\% on the Hybrid-SQuAD test set.
\end{abstract}

\section{Introduction}
\label{sec:introduction}

Question Answering (QA) systems take as input a natural language question and provide an answer from a predefined set of sources~\cite{zhang2023survey}. These sources may have structured data, as found in Knowledge Graphs (KGs), or unstructured data, such as text documents~\cite{dimitrakis2020survey}. 
To leverage the knowledge in both KG and text, hybrid QA has emerged~\cite{lehmann2024beyond}. Hybrid QA requires information from KG and text sources to generate the final answer~\cite{zhang2023survey}. The hybrid QA approach broadens the retrieval pieces of evidence across multiple sources, resulting in superior answer coverage compared to single-sourced QA models~\cite{feng-etal-2022-multi}. 

\begin{figure}[htb!]
  \centering
  \includegraphics[width=0.98\linewidth]{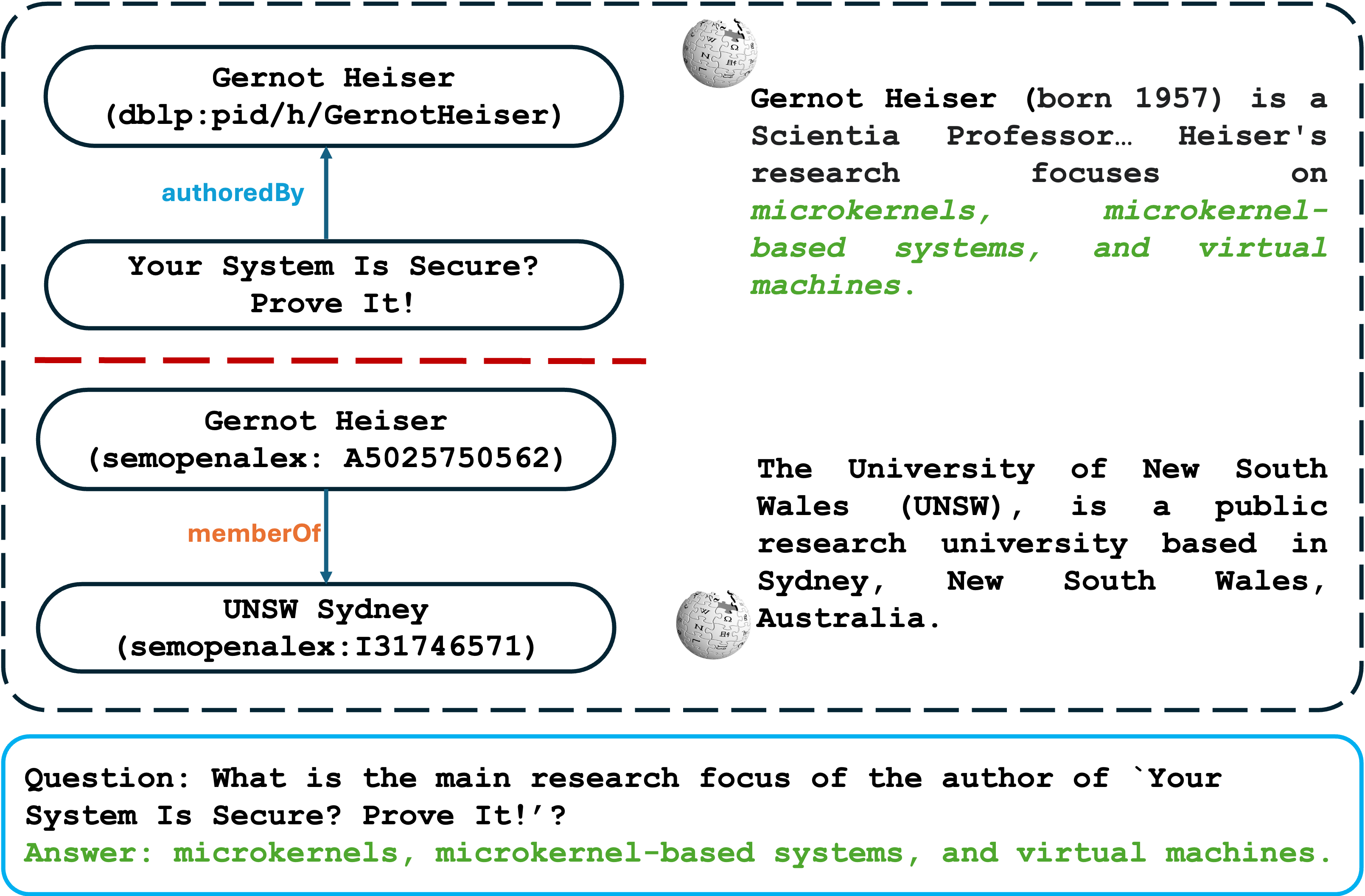}
  \caption{Sample scholarly information from heterogeneous data sources: DBLP, SemOpenAlex, and Wikipedia, along with a question and answer pair in Hybrid-SQuAD.}
  \label{fig:hetrogeneous_data_sources}
\end{figure}
In the context of Scholarly QA, existing models are designed to answer questions based solely on bibliographic metadata information found in a KG~\cite{DBLP:conf/semweb/TaffaU23,auer2023sciqa,jaradeh2020question} or in text~\cite{QASA10.5555/3618408.3619195,saikh2022scienceqa}. Additionally, the existing Scholarly QA data sets focus on dealing with homogeneous data, as listed in Table~\ref{tab:existing_scholarly_qa_dataset}. However, as shown in Figure~\ref{fig:hetrogeneous_data_sources}, information about scholarly entities like authors, publications, or institutions is distributed over heterogeneous sources. For instance, different facts about a scholar appear in the KGs - DBLP\footnote{\url{https://dblp.org}} (Digital Bibliography \& Library Project)~\cite{DBLP:conf/spire/Ley02} and SemOpenAlex\footnote{\url{https://semopenalex.org/resource/semopenalex:UniversalSearch}} (Semantic Open Alex)~\cite{farber2023semopenalex}, as well as in Wikipedia pages (See Figure~\ref{fig:hetrogeneous_data_sources}). Correspondingly, a question like ``What is the main research focus of the author of `Your System Is Secure? Prove It!'?" is only answerable by searching facts in both text and KG.

\begin{table*}[htb!]
  \centering
  \begin{tabularx}{0.95\textwidth}{p{4cm} p{1.5cm} p{1cm} X X}

  \hline
   Dataset&  Size & Multi-Source &  Source & Creation Method\\
   \hline
   DBLP-QuAD~\cite{DBLP:conf/birws/BanerjeeAUB23} &  10 K & no &  DBLP &  Template-based\\
   SciQA~\cite{auer2023sciqa} & 2.2 K & no & ORKG~\footnote{\url{https://orkg.org}} &  Template-based\\
   QASA~\cite{QASA10.5555/3618408.3619195} &  1.8 K & no &  Text & Human annotators\\
   \hline
   Hybrid-SQuAD (ours) &  10.5 K & yes &  DBLP, SemOpenAlex \& Text &  LLM-generated \\
   \hline
\end{tabularx}
\caption{Comparison of existing scholarly QA datasets}
\label{tab:existing_scholarly_qa_dataset}
\end{table*}

Therefore, to fill this gap and to foster the development of Scholarly hybrid QA, we create Hybrid-SQuAD (\textbf{Hybrid} \textbf{S}cholarly \textbf{Qu}estion \textbf{A}nswering \textbf{D}ataset) using an LLM (Large Languge Model). This new large-scale QA dataset requires looking at multiple data sources to provide an answer. Each question is aligned with DBLP and SemOpenAlex KG sub-graphs and Wikipedia text. 

Hybrid-SQuAD is particularly challenging for LLMs like ChatGPT-3.5\footnote{\url{https://openai.com/blog/chatgpt}}, which is trained on generic knowledge from the Web. While prompted with the questions from Hybrid-SQuAD, ChatGPT-3.5 can only achieve an accuracy of 2.6\% on the test set using its internal knowledge. In contrast, our baseline model demonstrates significantly higher performance, achieving a 69\% exact match score on the Hybrid-SQuAD test split.

The dataset is publicly available on our Github page~\url{https://github.com/semantic-systems/hybrid-squad}. 


\section{Related Works}

\subsection{Scholarly QA Datasets}

Common approaches in creating QA datasets are crowdsourcing, automatic generation, and collecting question-answer pairs from community-based QA platforms, such as Quora\footnote{\url{https://www.quora.com/}} and Stack Exchange\footnote{\url{https://stackexchange.com/}}~\cite{dzendzik_english_2021}. In crowdsourcing, from a given context, crowd-workers formulate question and answer pairs using in-house annotation tools or crowdsourcing annotation platforms like Amazon Mechanical Turk\footnote{\url{https://www.mturk.com/}}~\cite{chen-etal-2020-hybridqa}. On one hand, crowdsourcing allows for the creation of high-quality question-answer pairs, but it depends on the skill level of the workers. Crowdsourcing also typically incurs significant costs, especially for large-scale datasets. On the other hand, auto-generation approaches utilize language generation models, templates, or machine translation for the question-answer pairs formulation~\cite{dzendzik_english_2021}.

As shown in Table~\ref{tab:existing_scholarly_qa_dataset}, DBLP-QuAD, SciQA, and QASA are each derived from a single source. In contrast to these scholarly QA benchmarks, Hybrid-SQuAD introduces questions that require integrating structured knowledge from KGs with contextual understanding from text sources. Regarding their creation methods, DBLP-QuAD and SciQA use templates, while QASA relies on human annotators. Hybrid-SQuAD, however, employs an LLM to generate question-answer pairs. 

Conversely, CompMix~\cite{compMix2024} is a non-scholarly heterogeneous dataset that employs crowd-sourcing to generate question-answer pairs. This dataset capitalizes on the repetition of facts across sources like Wikipedia, Wikidata KG, Wikipedia tables, and info-boxes, allowing questions to be answerable by one or more of these underlying sources. Notably, the questions in the CompMix test set do not require reasoning across multiple sources. In contrast, our work highlights the complementary nature of three distinct sources: DBLP, SemOpenAlex, and Wikipedia text. For example, the KGs DBLP and SemOpenAlex lack personal details about authors, such as career milestones or institutional affiliations, which are exclusively found on Wikipedia pages dedicated to the authors and their institutions. On the other hand, KGs provide scholarly metadata like publication counts, citation numbers, and h-index metrics. Therefore, the three data sources used to construct Hybrid-SQuAD complement each other by offering a diverse array of information rather than reiterating the same data.

Furthermore, GPT-3~\cite{NEURIPS2020_1457c0d6} attains 50\% accuracy on the CompMix test set in a zero-shot setting, while ChatGPT-3.5 achieves only a 2.6\% exact match score on the test questions from Hybrid-SQuAD. This stark contrast highlights that Hybrid-SQuAD presents significantly more challenging questions, even for advanced LLMs.





\subsection{Scholarly Hybrid QA}

Contri(e)ve~\cite{shivashankar2024contrievecontextretrieve} presents a methodology that combines context extraction with prompt engineering. The context extraction process involves three stages: obtaining author information from the DBLP, retrieving data from SemOpenAlex using ORCID identifiers, and collecting additional details from Wikipedia. The collected data is refined to retain only relevant sentences with specific keywords, tackling issues associated with lengthy prompts that may degrade system performance. The subsequent prompt engineering phase organizes the prompts into four components—Instructions, Query, Context, and Output Indicator—ensuring they are concise yet informative for precise inference. Finally, the refined prompt is passed to the LLAMA3.1 8b-Instruct\footnote{\url{https://huggingface.co/meta-llama/Meta-Llama-3.1-8B-Instruct}} model to obtain the answer.

Similarly, Efeoglu et al.~\citeyear{sefika2024} extract triples from DBLP and SemOpenAlex, as well as relevant text from the Wikipedia corpus. They utilize an algorithm to create a context for each question by identifying pertinent triples and sentences. The relevance of these triples and sentences is determined through cosine similarity using SBERT\footnote{\url{https://sbert.net}} embeddings. This method allows for the selection of the most significant evidence by retaining only the essential components. The resulting evidence-matched context is then used to fine-tune the Flan-T5-Large\footnote{\url{https://huggingface.co/google/flan-t5-small}} model in a supervised setting, enabling it to effectively function as an answer extractor. 

Likewise, Fondi and Jiomekong~\citeyear{fondi2024integrating} collect data from all the three sources and structure it to facilitate pattern detection among queries through alphabetical ordering. They employ a divide-and-conquer strategy to group questions based on author identifiers and manually assigned topics. Ultimately, they generate context-specific predictions as answers using the BERT-base-cased-squad2\footnote{\url{https://huggingface.co/deepset/bert-base-cased-squad2}} language model.

Contrary to these methodologies, we identify sub-question phrases, resolve the scholarly entities they contain, and replace each phrase with its corresponding entity. Subsequently search relevant evidence in KGs and textual sources. Finally, utilize a RAG (Retrieval-Augmented Generation)~\cite{rag_10.5555/3495724.3496517} model to generate an answer.



\section{The Dataset}


This section outlines the process for collecting triples from the KGs along with their associated passages, followed by a description of the question generation method. Additionally, it provides various statistics related to Hybrid-SQuAD.

\subsection{Data Collection}

Our process for creating questions across multiple data sources starts by downloading the DBLP\footnote{\url{https://blog.dblp.org/2022/03/02/dblp-in-rdf/}} RDF dump. This choice is based on DBLP's focus on Computer Science publications, its manageable size, its well-defined schema, and the inclusion of Wikipedia URLs for authors, simplifying text retrieval from Wikipedia. We also use SemOpenAlex~\cite{farber2023semopenalex}, specifically downloading only the Authors\footnote{\url{https://semopenalex.org/authors/context}} and Institutions\footnote{\url{https://semopenalex.org/institutions/context}} dump files due to the large size of the Publications file. Although SemOpenAlex lacks authors' Wikidata IDs or Wikipedia URLs, making Wikipedia page retrieval challenging, it provides valuable statistical information that complements DBLP data. We crawl Wikipedia to collect textual data on authors and their institutions using the URLs from DBLP and SemOpenAlex, ensuring a comprehensive dataset. 
The data collection steps are:

\noindent\textbf{Step 1: Retrieval of DBLP Authors and their Information}
\begin{itemize}
\item Retrieve authors who have a Wikipedia URI and an ORCID\footnote{ORCID is used to link an author in DBLP with its corresponding author in SemOpenAlex.} (Open Researcher and Contributor ID)\footnote{\url{https://orcid.org}}, along with their names and primary affiliations (see SPARQL-1 in Appendix~\ref{sec:appendixA}). 
\item For each author extract publications (see SPARQL-2 in Appendix~\ref{sec:appendixA}).
\item Use the authors' Wikipedia\_uri to extract the corresponding Wikipedia text, then clean the text by utilizing an HTML and XML parser Python library -  BeautifulSoup\footnote{\url{https://pypi.org/project/beautifulsoup4/}} to remove HTML tags, references, extra spaces, and links.
\end{itemize}

\noindent\textbf{Step 2: Searching and Extraction of Corresponding Authors and Institutions information from SemOpenAlex}
\begin{itemize}
\item Identify DBLP author's corresponding SemOpenAlex URI, using ORCID as a matching criterion (see SPARQL-3 in Appendix~\ref{sec:appendixA}).
\item Extract the author's hIndex, I10Index, number of publications, citations, and two-year average citedness from SemOpenAlex (see SPARQL-4 in Appendix~\ref{sec:appendixA}).
\item Collect the author's institution number of publications, number of citations, institution type, and institution Wikipedia URI from SemOpenAlex (see SPARQL-5 in Appendix~\ref{sec:appendixA}).
\item Extract the author's institution Wikipedia text through the provided URI and clean up the text using BeautifulSoup.
\end{itemize}

Finally, we compile a comprehensive data source pool from the three sources into a JSON file.

\subsection{Question-Answer Pair Generation}


The question-answer pair generation involves three main steps: context preparation, prompt construction, and generation.
\begin{mdframed}
\captionof{lstlisting}{Question Generation Prompt.}
\label{question-generation-prompt_template}
\begin{verbatim}
Task: You are an experienced annotator.
Your task is to formulate fact-seeking
questions from the given data.  
Hint: Follow the given Example.
Do not add anything else.
Note: Ensure the answer is a word or 
set of consecutive words in the source.
Example: 
Source 1: {sample-data1}
Source 2: {sample-data2}
{examples}
Source 1: {source1-data}
Source 2: {source2-data}
question: 
answer:
\end{verbatim}
\end{mdframed}
\subsubsection{Context Preparation} 
For bridging questions over both DBLP \& SemOpenAlex (KG-KG) or a combination of a KG and text (KG-Text), a record of an author is selected from the pool, along with its DBLP facts (as source 1) and the corresponding SemOpenAlex facts (as source 2). In the case of KG-Text questions, the first source is taken from DBLP or SemOpenAlex records, and the textual information of the DBLP entity taken from Wikipedia becomes the second source. Unlike bridging questions, for KG-KG comparison questions, 
, the data source pool is split into two distinct lists. A DBLP entity and its corresponding SemOpenAlex facts are chosen from one half and set as the first source. Another entity DBLP record is selected from the other half, along with SemOpenAlex data, and assigned as a second source. The first source for questions involving KG-KG-Text inference is an entity's DBLP records. The second source comprises the corresponding SemOpenAlex records, and the third is the textual information.

\begin{table}[htb!]
    \centering
    \begin{tabular}{ll}
    \hline Category & Count\\ \hline
  Bibliometric Numbers & 31 \\
  Biographical Information & 21 \\
  Organization & 19 \\ 
  Location & 11 \\ 
  Date & 8 \\ 
  Research Works & 7 \\ 
  Other & 7 \\ \hline
    \end{tabular}
    \caption{Answer Categories of randomly selected 100 questions in the test set.}
    \label{tab:questions_expected_answer_types}
\end{table}

\begin{table*}[htb!]
  \centering
  \begin{tabularx}{0.95\textwidth}{p{5cm} p{1.5cm} X}
   \hline
   \textbf{Types with Examples} & Percentage (\%) & Description\\ \hline
Type I \newline KG -> KG (Bridge)\newline Q: What is the citation of the author who collaborated with Piero Fraternali on `The Story of the IDEA Methodology'?
\newline A: 14182 & 23.93 & Necessitates traversing the connections within KGs to identify relevant data points. For instance, to ascertain an author's identity, one starts with the author’s information in DBLP, then navigates to SemOpenAlex to resolve the entity—an author or an institution—through the SemOpenAlex author URI. The final answer emerges from the consolidated information deduced from these interconnected KGs, such as identifying the citations of institutions.\\ \hline
Type II \newline KG -> KG (Comparison) \newline Q: Whose twoYearMeanCitedness is greater, Joshua R. Smith or Duncan J. Watts? \newline A: Duncan J. Watts & 9.97 & Focus on comparing attributes across different entities residing in separate KGs. An example of this involves determining the author with the higher hIndex. The question requires retrieving an author’s details from DBLP and accessing their corresponding hIndex data from SemOpenAlex. \\ \hline
Type III \newline KG -> Text \newline Q: What is the main research focus of the author of `Your System Is Secure? Prove It!'? \newline A: microkernels, microkernel-based systems, and virtual machines & 55.55 & Such pathways require first identifying an entity based on its properties from a KG. This is followed by locating a text snippet that answers the question. For instance, to answer ``What is the main research focus of the author of `Your System Is Secure? Prove It!'?" one must initially retrieve the author's name, Gernot Heiser, from DBLP. Subsequently, a text source is needed to determine the author's primary research themes.\\ \hline
Type IV \newline KG -> KG -> Text \newline Q: What is the motto of the academic institution where Russell Greiner is affiliated? \newline A: Quaecumque vera (whatsoever things are true) & 10.54 & Require navigating evidence from two KGs—specifically, identifying the author and their institution via DBLP and SemOpenAlex. Then, extracting information about the motto from the institution's Wikipedia entry will yield the answer.\\ \hline
\end{tabularx}
\caption{Distribution of traversal pathways in Hybrid-SQuAD test set questions, highlighting the percentage of each pathway utilized with their description.}
\label{tab:reasoning_types}
\end{table*}

\subsubsection{Question Generation} 
In the question generation, we first construct prompts that include toy data, instructions, examples, and data about an entity from two sources (see Listing~\ref{question-generation-prompt_template}). We then use ChatGPT-3.5 to process the prompts and format the generated question-answer pairs into JSON format. During a random check on the question-answer pairs, we found instances where some questions either lacked an answer or had responses consisting of only a single letter. The issue arose because we initially provided the entire text and the entity KG facts to the LLM in one go, resulting in truncated answers. We re-prompt the LLM to address this by supplying the question and its context. Then, replace the incomplete answers with these new, more complete responses. Finally, as shown in Appendix~\ref{sec:appendixB}, a single comprehensive JSON file compiles all 10,581 generated question-answer pairs and their unique IDs, Author DBLP URIs, source types, and question types.



\section{Dataset Analysis}

\subsection{Answer Types}
Our manual analysis of 100 randomly selected questions, presented in Table~\ref{tab:questions_expected_answer_types}, revealed that 31 answers fall under the `bibliometric numbers' category. These questions typically inquire about bibliometric metrics, including publication counts, citation numbers, the h-index, and the i10-index. The category of biographical information encompasses 21 answers that provide insight into a scholar's educational background, professional experience, and academic achievements. Furthermore, 19 questions are classified under the category of `Organization', which includes the names of Universities, research centers, or laboratories. Answers about places are classified under the category of `location' (11 questions). In comparison, those providing specific dates (8 questions) offer contextual information such as the establishment year of an institute or dates of notable scholarly events.
Additionally, questions about research outputs and publications are placed under the `research works' category, including seven questions highlighting inquiries into scholarly contributions. The remaining seven questions that do not fit into these predefined categories are grouped under the `other' category, indicating the inquiries' diverse nature. Besides, Figure~\ref{fig:question_term_distribution} visualize the first consecutive four words in the questions.

\subsection{Question Evidence Traversal Paths}

Table~\ref{tab:reasoning_types} analyzes the 700 test set questions in Hybrid-SQuAD, highlighting four paths for evidence traversal used in the test set questions. The analysis shows a significant reliance on the KG -> Text pathway, utilized in 55.55\% of the questions. This highlights the need to effectively integrate KG data with textual information to generate accurate answers. The distribution of the remaining questions emphasizes the direct and comparative evidence traversal: KG -> KG (Bridge) accounts for 23.93\%, KG -> KG -> Text for 10.54\%, and KG -> KG (Comparison) for 9.97\%. This distribution illustrates how most questions leverage structured and unstructured data, combining elements from KGs and texts to formulate comprehensive answers.

\begin{figure*}[htb!]
  \centering
  \includegraphics[width=0.95\textwidth]{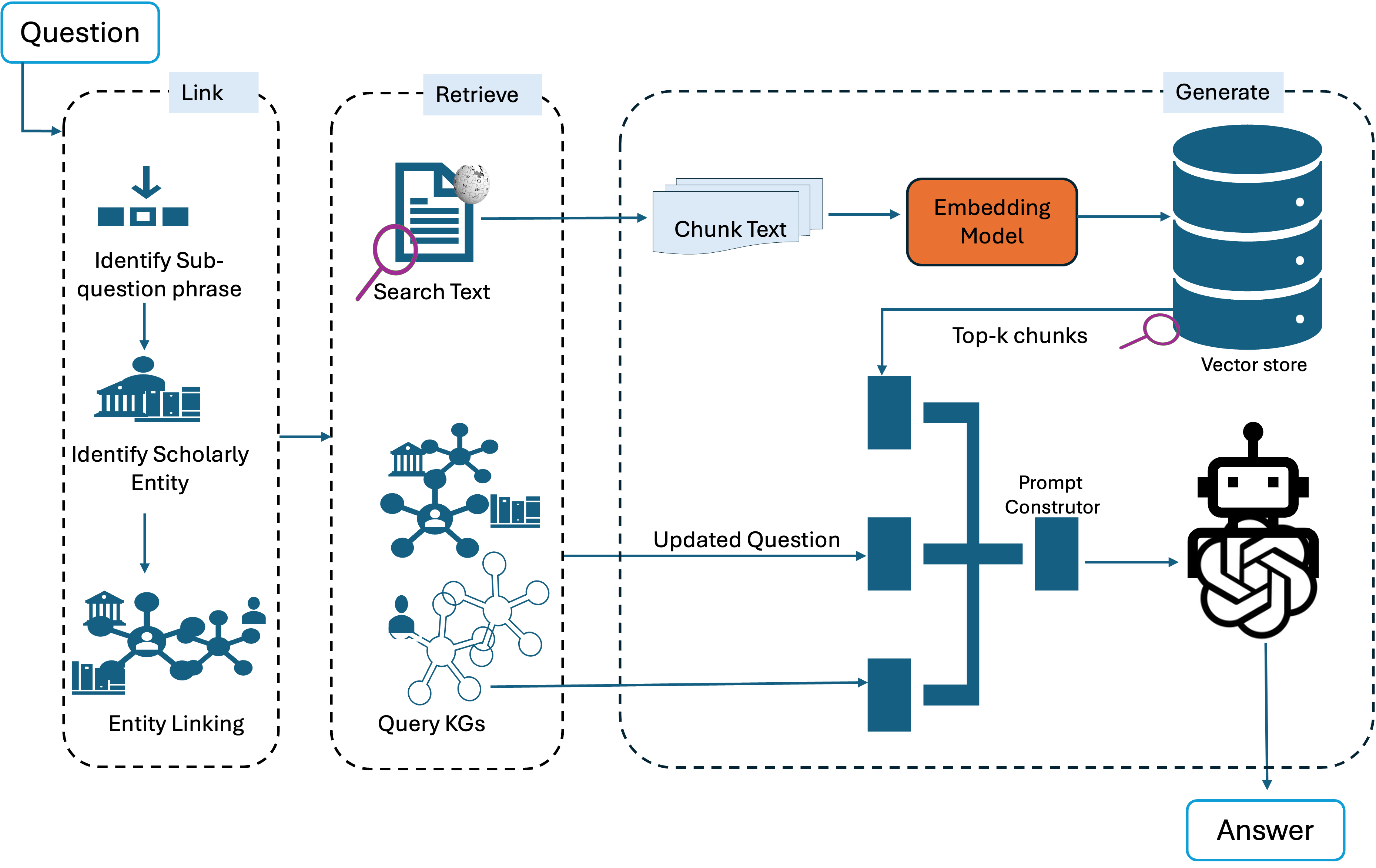}
  \caption{Baseline Model}
  \label{fig:baseline_model}
\end{figure*}

\section{Baseline Model}

As shown in Figure~\ref{fig:baseline_model}, the baseline model for answering questions comprises three phases: link, retrieve, and generate. The model identifies the bridging entity that connects various data sources in the link phase. This ensures that the different pieces of information are properly aligned by recognizing scholarly entities such as publications, author names, or institutions. Once a bridging entity is identified, the retriever module searches for relevant data about an entity from text sources and KGs. The retrieval phase allows comprehensive information from KG and text to be gathered. Finally, in the generate phase, the model fuses the heterogeneous inputs obtained from the retriever in the LLM and generates an answer. 

\subsection{Linking}

The questions in Hybrid-SQuAD necessitate searching for evidence across multiple data sources and often include embedded sub-questions within the main query. For example, in the question, ``What is the main research focus of \textit{the author of `Your System Is Secure? Prove It!’}?”, the phrase in italics represents the sub-question. Answering such queries requires the identification of these sub-question phrases and the scholarly entities involved, such as the publication titled \textit{`Your System Is Secure? Prove It!’} and the resolution of the author entity. These sub-question phrases and entities bridge different data sources, facilitating a comprehensive understanding and integration of information.

The sub-question phrase identification sub-component forms a prompt with a few examples and then prompts ChatGPT-3.5 to extract the sub-question. Subsequently, the scholarly entity identifier prompts the LLM to identify scholarly entities within the sub-question phrase, such as a publication, author, or institution. Once these scholarly entities are identified, the entity linker employs a SPARQL template to query DBLP, retrieving the relevant URLs and labels for these entities. If the entity linker successfully determines the label of the bridging entity, it updates the question by replacing the sub-question phrase with the resolved entity. This process is repeated recursively if multiple sub-question phrases exist, ensuring the question is thoroughly updated. The updated question, accompanied by the entity labels, is then forwarded to the retriever component. When the sub-question phrase and entity are not identifiable, the entity linker uses the author URI provided with the question to resolve the bridging entity. This rigorous linking process guarantees that the question is accurately transformed and enriched with precise entity information before being passed on to the retriever for context extraction.

\subsection{Retrieve}
The retrieval stage integrates both text search and KG query functionalities. When the evidence pathway involves KG-Text or KG-KG-Text sequences, the process begins with a text search. This search consists of extracting Wikipedia text related to the author or institution entity by utilizing the Wikipedia URL provided by the entity linker. The extracted text is parsed using BeautifulSoup, resulting in a plain text excerpt. In situations requiring a KG-KG, the KG query sub-component uses a SPARQL template from SemOpenAlex to gather all pertinent facts about the entity. This is achieved by leveraging the author’s ORCID\footnote{\url{https://orcid.org}} as a linkage criterion between DBLP and SemOpenAlex. The gathered texts, KG details, and the updated query are ultimately fed into the answer generation stage to produce the final response.

\subsection{Generate}

The answer generation process for questions involving KG-Text or KG-KG-Text evidence pathways leverages RAG, combining document retrieval with LLM to ensure precise responses. The RAG model loads the text, divides it into 200-word chunks with a 10-word overlap, generates embeddings using the `BAAI/bge-small-en-v1.5’\footnote{\url{https://huggingface.co/BAAI/bge-small-en-v1.5}} model, and stores these embeddings in a FAISS~\cite{douze2024faiss} vector store for efficient retrieval. The FAISS vector store retrieves top-5 relevant text chunks using the updated question as a query. The ChatGPT-3.5 model then processes these chunks alongside the KG triple labels to generate an answer. In contrast, for KG-KG evidence pathways, the LLM only receives the updated question and KG triple labels. 

\section{Evaluation}

To evaluate the performance of our baseline model, we use Exact Match (EM) evaluation metrics that assess the proportion of predictions that match the gold answers, reflecting the model's ability to produce precise outputs. We also use F-Score evaluation metrics.

\begin{table}
    \centering
    \begin{tabular}{lll}
    \hline
        Models &  EM  & F-Score \\ \hline 
        Fondi and Fidel~\citeyear{fondi2024integrating} & 43.59 & 45.05 \\
         Contri(e)ve~\citeyear{shivashankar2024contrievecontextretrieve} & 32.0 &	40.7 \\
        Efeoglu et al.~\citeyear{sefika2024} & 48.9 & - \\\hline
        Ours \\ \hline
        RAG-based (ChatGPT-3.5) & \textbf{69.65} & \textbf{74.91} \\
        RAG-based (LLAMA-3-8B) & 61.1 & 68.92 \\
    \hline
    \end{tabular}
    \caption{Exact Match score of different models tested on Hybrid-SQuAD and Our RAG Based Baseline Model using ChatGPT-3.5 and LLAMA-3-8B as Generator.}
    \label{tab:evaluation_result}
\end{table}
As shown in Table~\ref{tab:evaluation_result}, the fine-tuned Flan-T5 Large by~\cite{sefika2024}\footnote{Evaluated on a different test split.} and Contri(e)ve~\cite{shivashankar2024contrievecontextretrieve} achieve an EM score of 48.9 and 32.0, respectively. In contrast, our baseline models demonstrate superior performance, with the RAG-based approach using ChatGPT-3.5 achieving the highest EM score of 69.65 and 74.91 F-score, underscoring the effectiveness of RAG strategies in enhancing accuracy through contextual information. The RAG-based model using LLAMA-3-8B also performs well with an EM and F-score of 61.1 \& 68.92 respectively, significantly outperforming the other models and reinforcing the benefits of integrating retrieval mechanisms into language model frameworks.

\begin{table}
    \centering
    \begin{tabular}{lll}
    \hline
        Model &  EM  & F-Score \\ \hline  
        Zero-shot (ChatGPT-3.5) & 2.6 & 8.91 \\
        Zero-shot (LLAMA3.0-8B) &  1.3 &  8.00 \\
    \hline
    \end{tabular}
    \caption{Zero-shot Exact Match scores of ChatGPT-3.5 and LLAMA-3-8B on Hybrid-SQuAD.}
    \label{tab:chatgpt_evaluation_result}
\end{table}
To see how the LLMs behave on Hybrid-SQuAD, we have prompted ChatGPT and LLAMA in Zero-shot without any contextual input, and as Table~\ref{tab:chatgpt_evaluation_result} shows, the performance is very minimal. The Zero-shot using ChatGPT-3.5 Turbo achieved an exact match score of 2.6, while LLAMA3.0-8B performed even less effectively, with an EM score of 1.3. These LLMs are trained primarily on generic domain text from the web; they need help with questions requiring heterogeneous scholarly information typically found in scholarly KGs. The findings emphasize the importance of integrating supplementary retrieval mechanisms to improve LLMs' ability to effectively address complex questions, especially those involving domain-specific knowledge. 

\section{Summary}

This paper introduces Hybrid-SQuAD, a large-scale hybrid Scholarly QA dataset. Hybrid-SQuAD contains questions that need multiple data sources to be able to answer. Current LLMs, such as ChatGPT-3.5, perform poorly on this dataset, with results in the range of 3\% accuracy, while a baseline QA system achieves 69.65 exact match and 74.91 F-Score. Additionally, hope our novel benchmark sparks further research on Scholarly hybrid QA. 



\bibliography{references}
\appendix
\section{Appendix A}
\label{sec:appendixA}
\begin{verbatim}
# SPARQL-1
PREFIX rdf: <http://www.w3.org/1999/02/
             22-rdf-syntax-ns#>
PREFIX dblp: <https://dblp.org/rdf/schema#>
SELECT * WHERE {
  ?auth_dblp_uri dblp:wikipedia ?wikipedia ;
    rdf:type dblp:Person ;
    dblp:orcid ?orcid ;
    dblp:creatorName ?name ;
        dblp:primaryAffiliation ?affiliation.
}

# SPARQL-2
PREFIX dblp: <https://dblp.org/rdf/schema#>
SELECT DISTINCT *
WHERE {
%s ^dblp:authoredBy ?publication .
?publication dblp:title ?title ;
    dblp:publishedIn ?publishedin ;
    dblp:yearOfPublication ?year ;
dblp:numberOfCreators ?numberOfCreators .
FILTER (?numberOfCreators < 3)
 }
ORDER BY DESC(?year)

# SPARQL-3
PREFIX ns2: <https://semopenalex.org/ontology/>
PREFIX ns3: <http://purl.org/spar/bido/>
PREFIX ns4: <https://dbpedia.org/ontology/>
PREFIX ns5: <https://dbpedia.org/property/>
SELECT * WHERE
{
   OPTIONAL {?auth_soa_uri ns2:orcidId ?orcid .}
   OPTIONAL {?auth_soa_uri ns3:orcidId ?orcid .}
   OPTIONAL { ?auth_soa_uri ns4:orcidId ?orcid .}
   OPTIONAL { ?auth_soa_uri ns5:orcidId ?orcid .}
   FILTER (?orcid = "%s")
}
# s is a variable representing orcid.

# SPARQL-4 
PREFIX soa: <https://semopenalex.org/ontology/>
PREFIX foaf: <http://xmlns.com/foaf/0.1/>
PREFIX org: <http://www.w3.org/ns/org#>
PREFIX ns3: <http://purl.org/spar/bido/>
SELECT *
WHERE {
%s foaf:name ?author_name ;
   soa:worksCount ?worksCount ;
   soa:citedByCount ?citedByCount ;                        
   ns3:h-index ?hIndex ;
   soa:i10Index ?i10Index ;
   soa:2YrMeanCitedness ?twoYearMeanCitedness .
}
# s is a placeholder for soa_author_uri.

# SPARQL-5
PREFIX rdfs: <http://www.w3.org/
             2000/01/rdf-schema#>
PREFIX owl: <http://www.w3.org/2002/07/owl#>
PREFIX soa: <https://semopenalex.org/ontology/>
PREFIX foaf: <http://xmlns.com/foaf/0.1/>
PREFIX ns5: <https://dbpedia.org/property/>
SELECT *
WHERE {
%s foaf:name ?institute_name ;
    soa:worksCount ?publicationsCount ;
    soa:citedByCount ?publicationsCitedByCount ;
    soa:rorType ?institute_type ;
    ns5:countryCode ?institute_country_code ;
    rdfs:seeAlso ?wikipedia_url .
FILTER (CONTAINS(STR(?wikipedia_url), 
        "en.wikipedia.org"))
}
# s is a placeholder for soa_author_uri.
\end{verbatim}

\section{Appendix B}
\label{sec:appendixB}
\begin{mdframed}
\captionof{lstlisting}{Sample Generated Question.}
\label{sample-question-in-json}
\begin{verbatim}
[{
"auth_dblp_uri": 
      "<.../pid/c/StefanoCeri>",
"id": "...-cc4875320424",
"question": "What is the cited by count 
of the author who collaborated with 
Piero Fraternali on 
'The Story of the IDEA Methodology'?",
"answer": "14182",
"type": "bridge",
"source_types": ["KG", "KG"]
},...]
\end{verbatim}
\end{mdframed}

\begin{figure*}[htb!]
  \centering
  \includegraphics[width=0.9\textwidth]{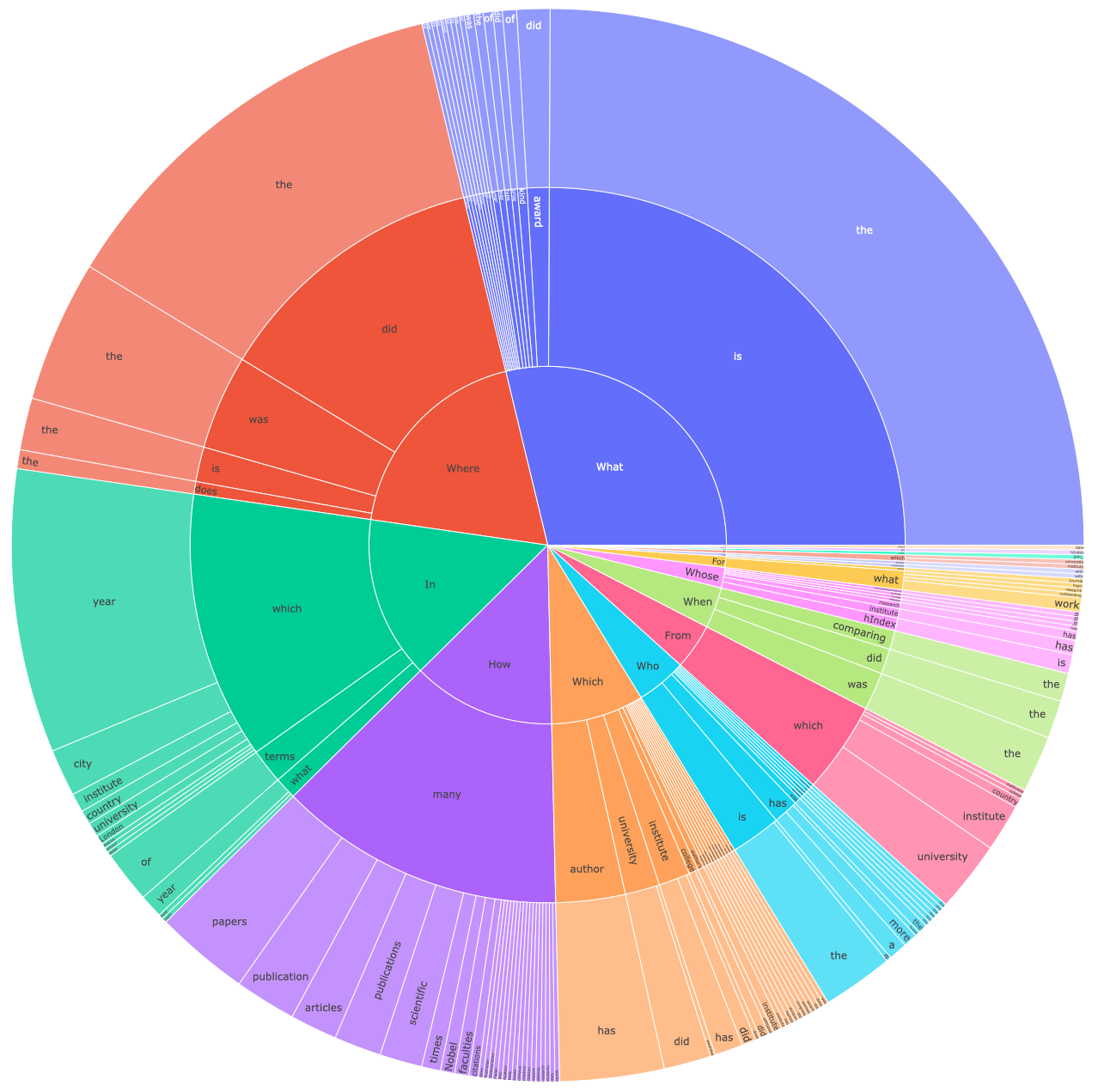}
  \caption{First three words distributions in questions.}
  \label{fig:question_term_distribution}
\end{figure*}

\end{document}